\documentclass[letterpaper, 10 pt, conference]{ieeeconf}  
\usepackage{graphicx}
\usepackage{pdfpages}
\usepackage{array} 
\usepackage{nameref}
\usepackage{multirow} 
\usepackage{booktabs} 
\usepackage{hyperref}
\usepackage{algorithm}
\usepackage{algorithmic}
\usepackage{xcolor} 
\usepackage{tabularx}
\usepackage{threeparttable}

\IEEEoverridecommandlockouts                              

\usepackage{amsmath} 
\usepackage{amssymb}  

\title{\LARGE \bf
IMM-MOT: A Novel 3D Multi-object Tracking Framework with Interacting Multiple Model Filter
}

\author{Xiaohong Liu$^{\dagger}$, Xulong Zhao$^{\dagger}$, Gang Liu$^{*}$, Zili Wu, Tao Wang, Lei Meng, Yuhan Wang
\thanks{$^{\dagger}$ These authors contributed equally to this work.}
\thanks{$^*$ Corresponding author.}
\thanks{All authors are with School of Computer Science and Technology, Xidian University,  Xi’an, 710126, Shaanxi, China.}
}

\begin{document}

\maketitle
\thispagestyle{empty}
\pagestyle{empty}

\begin{abstract}
3D Multi-Object Tracking (MOT) provides the trajectories of surrounding objects, assisting robots or vehicles in smarter path planning and obstacle avoidance. Existing 3D MOT methods based on the Tracking-by-Detection framework typically use a single motion model to track an object throughout its entire tracking process. However, objects may change their motion patterns due to variations in the surrounding environment. In this paper, we introduce the Interacting Multiple Model  filter in IMM-MOT, which accurately fits the complex motion patterns of individual objects, overcoming the limitation of single-model tracking in existing approaches. In addition, we incorporate a Damping Window mechanism into the trajectory lifecycle management, leveraging the continuous association status of trajectories to control their creation and termination, reducing the occurrence of overlooked low-confidence true targets. Furthermore, we propose the Distance-Based Score Enhancement  module, which enhances the differentiation between false positives and true positives by adjusting detection scores, thereby improving the effectiveness of the Score Filter.  On the NuScenes Val dataset, IMM-MOT outperforms most other single-modal models using 3D point clouds, achieving an AMOTA of 73.8\%. Our project is available at https://github.com/Ap01lo/IMM-MOT.
\end{abstract}

\section{INTRODUCTION}
3D Multi-Object Tracking (MOT) aims to generate accurate temporal trajectories for each target while maintaining consistent identities, even as targets move or scenes change. By providing target trajectories, 3D MOT facilitates intelligent path planning or target following for robots or vehicles, which is crucial in robotic perception and autonomous driving.
Currently, 3D MOT primarily adopts two frameworks, “Tracking by Detections” (TBD) \cite{weng3DMultiObjectTracking2020,nagyRobMOTRobust3D2024} and “Joint Detection and Tracking” (JDT) \cite{240313443FastPolyFast,huangJointMultiObjectDetection2021,ahmarEnhancingThermalMOT2024}. TBD stands out for its exceptional accuracy and robustness. Recent state-of-the-art methods in 3D MOT, including MCTrack \cite{wangMCTrackUnified3D2024}, FastPoly \cite{240313443FastPolyFast}, BoostTrack \cite{stanojevicBoostTrackUsingTracklet2024}, and shaSTA \cite{sadjadpourShaSTAModelingShape2024}, all adopt the TBD framework.

\begin{figure}
\centering
\resizebox{\columnwidth}{!}{\includegraphics{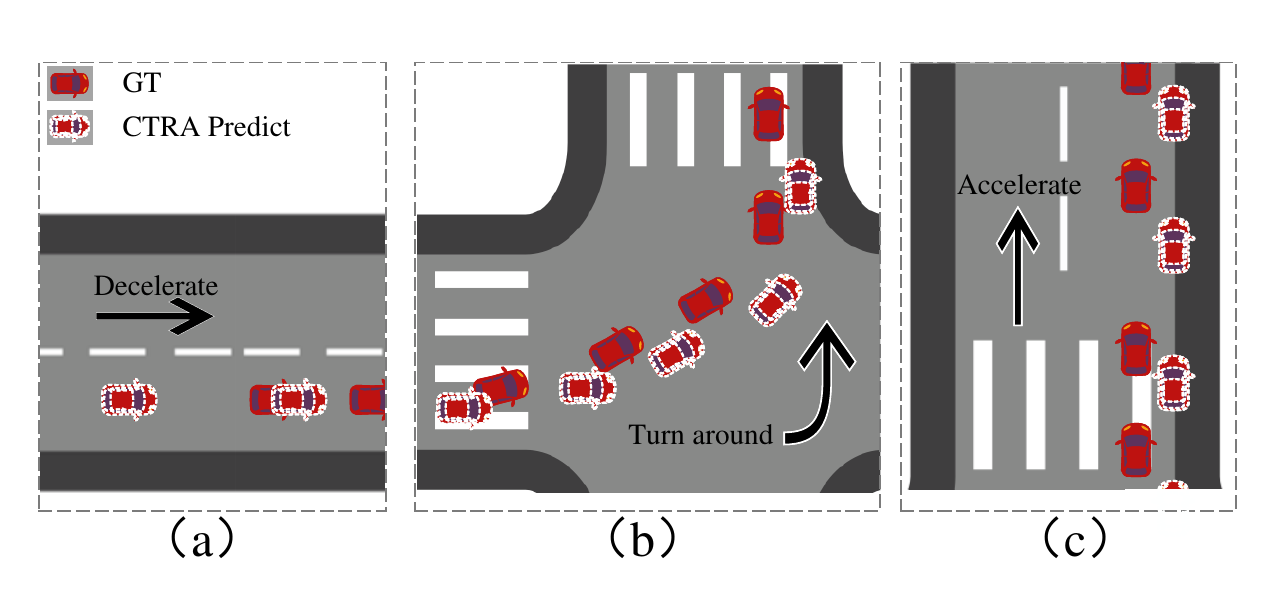}}

\caption{An illustration of a car turning left and its sequential scenarios: (a) Deceleration; (b) Turning; (c) Acceleration.}
\label{fig1}
\end{figure}

Despite the superior performance of TBD-based methods, unresolved issues remain in various modules. 
(1) The preprocessing module often uses Non-Maximum Suppression (NMS) and Score Filter (SF) with fixed confidence thresholds to filter detections. However, such an absolute filtering method directly eliminates all detections below the threshold, which leads to the misjudgment of true targets, an increase in false negatives (FN), and reduced tracking quality.
(2) In the state prediction stage, filter-based tracking methods \cite{weng3DMultiObjectTracking2020,anhEnhancedKalmanAdaptive2024,caoPKFProbabilisticData2024} primarily use corresponding motion models to predict the states of targets. Poly-MOT \cite{liPolyMOTPolyhedralFramework2023} observed that using a single motion model to predict targets of different categories resulted in low accuracy \cite{weng3DMultiObjectTracking2020,kimEagerMOT3DMultiObject2021}. Poly-MOT proposed employing different motion models for various target types to address this problem. Although this approach improved tracking performance, it overlooked the fact that the motion patterns of the same target type may change during different motion stages. For instance, as shown in Fig. \ref{fig1}, a car might accelerate, then turn, and subsequently decelerate. Using a single motion model to fit such motion states is not reasonable.
(3) The trajectory management stage commonly used methods include count-based \cite{liPolyMOTPolyhedralFramework2023} and confidence-based \cite{benbarkaScoreRefinementConfidencebased2021} approaches. The count-based approach updates trajectory states by calculating the consecutive times a trajectory is associated or dissociated. However, due to its simplicity, the adjustment of thresholds significantly impacts results, making it challenging to balance the number of false positives (FP) and FN \cite{benbarkaScoreRefinementConfidencebased2021}. The confidence-based approach adjusts states through a score update function. However, its score decay mechanism often neglects low-confidence targets, resulting in missed detections.

Thus, we propose IMM-MOT, a comprehensive 3D MOT framework, based on the baseline Poly-MOT \cite{liPolyMOTPolyhedralFramework2023}. In the tracking phase, we introduce  an Interacting Multiple Model (IMM) filter to track targets. The challenge of varying motion patterns for the same target across different motion stages is addressed by combining multiple motion models. We propose a trajectory management mechanism based on damping window scores in the trajectory management phase, using the association states of trajectories as the basis for trajectory initiation and termination, effectively reducing the number of  FN. Additionally, we introduce a Distance-Based Score Enhancement (DBSE) method to improve the separation between true positives (TP) and FP, enhancing the performance of SF.  
Our main contributions include:  
\begin{itemize}  
    \item A tracker based on the IMM algorithm is proposed to rapidly adapts to target motion patterns. Our method provides more accurate target predictions for the association phase by dynamically fitting a target's motion using multiple models during its motion process. 
    \item We propose a trajectory management mechanism based on a Damping Window (DW) mechanism , which incorporates the association states of all detections within a trajectory. It then uses a damping function to compute trajectory scores for state updates, not only accounting for low-quality true detections but also improving tracking quality more effectively. 
    \item For LiDAR point cloud detectors, the DBSE mechanism is proposed. This mechanism employs an enhancement function that leverages the density characteristics of LiDAR point clouds, increasing the scores of nearby targets with high confidence and reducing the scores of distant targets with low confidence. This approach not only amplifies the distinction between true and false detections in terms of detection scores but also strengthens SF's ability to filter FP.  
    \item Our IMM-MOT framework achieves an AMOTA of 73.8\% on the NuScenes Val Set, which is the highest known value for methods using the CenterPoint point cloud detector.  
\end{itemize}  

\begin{figure*}[t]
  \centering
  \resizebox{\linewidth}{!}{\includegraphics{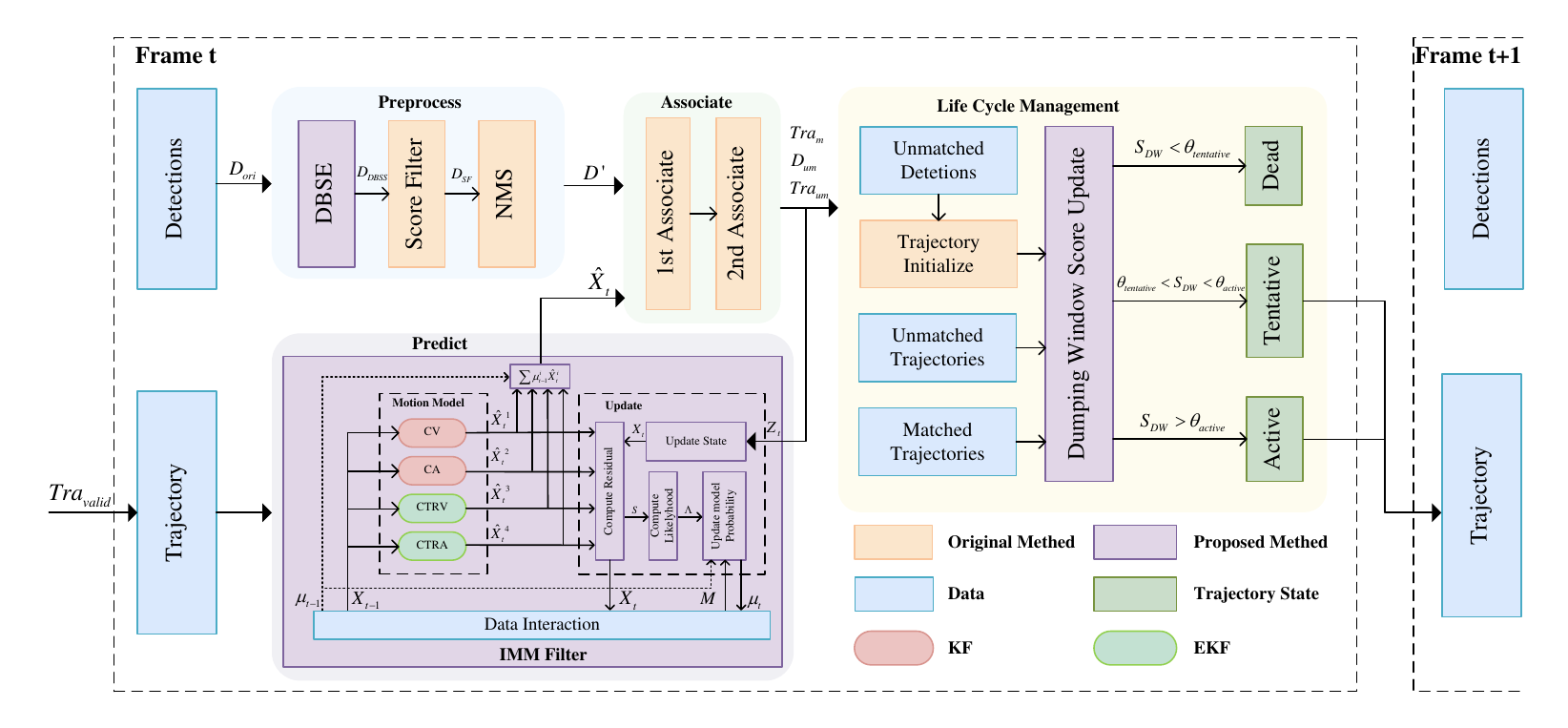}}

  \caption{The framework of IMM-MOT is as follows:  
(1) Distance-Based Score Enhancement (DBSE), Score Filter (SF), and Non-Maximum Suppression (NMS) are applied sequentially to process the detection values in the preprocessing stage, resulting in $D'$. $D'$ is then fed into the association module.  (2) The Interacting Multiple Model (IMM) tracker predicts the next-frame target positions of effective trajectories in the trajectory library, resulting in $\hat{X}_t$. Each model predicts based on the previous frame’s $X_{t-1}$, and the predicted values $\hat{X}_t^i$ from each model are weighted by the previous frame’s model probabilities $\mu_{t-1}^i$ to obtain $\hat{X}_t$. $\hat{X}_t$ is then passed into the association module to obtain the association results. The observations $Z_t$ from successful associations are used to update the model probabilities. During the update, residuals $S_t^i$ are first calculated, followed by the computation of the likelihoods $\Lambda_t^i$ for each model. Combining these with the previous frame’s $\mu_{t-1}$ and the Markov matrix M, the updated model probabilities $\mu_t$ are obtained.
(3) The trajectory initialization process determines whether new tracks are formed for unassociated detection values after association. The Damping Window (DW) scores $S_{DW}$ for all trajectories are computed and compared with the active and tentative thresholds $\theta_{active}$ and $\theta_{tentative}$, respectively, to determine the trajectory states. Active and tentative trajectories are retained as effective trajectories for tracking in subsequent frames.}
  \label{fig3}
\end{figure*}

\section{Related Work}

\textbf{3D Multi-Object Tracking.} The ongoing advancements in 3D object detection \cite{chenLargeKernel3DScalingKernels2023,yinCenterbased3DObject2021,xiePolyPCPolyhedralNetwork2023} and 2D MOT \cite{wojkeSimpleOnlineRealtime2017,zhang2022bytetrack} technologies have significantly propelled innovation and progress in the field of 3D MOT. AB3DMOT \cite{weng3DMultiObjectTracking2020} first employs a Kalman filter for prediction and uses “Intersection over Union” (IoU) association, while also introducing a new evaluation tool for 3D MOT. SimpleTrack \cite{pangSimpleTrackUnderstandingRethinking2021} modularizes the TBD-based 3D MOT method into four components and proposes a two-stage association strategy. To address the issue of multi-class targets during tracking, Poly-MOT \cite{liPolyMOTPolyhedralFramework2023} introduces the first multi-class tracking framework, incorporating geometric constraints into the motion model. Building on Poly-MOT, Fast-Poly \cite{240313443FastPolyFast} introduces the concept of A-GIoU and a lightweight filter, which improves the FPS. Although single-modal methods rely on a single data source, they tend to have relatively simple systems and low computational costs. Moreover, recent single-modal methods like Poly-MOT and Fast-Poly have outperformed many multi-modal approaches, highlighting their significant potential for further development.

\textbf{Preprocess.} Preprocessing mainly addresses two issues: (1) Duplicate detections for individual objects: Non-Maximum Suppression (NMS) removes duplicate detections based on the overlap between bounding boxes, with methods such as  Soft-NMS \cite{bodlaSoftNMSImprovingObject2017} and Adaptive-NMS \cite{liuAdaptiveNMSRefining2019} being commonly used. (2) Low-quality false detections: Score Filter (SF) filters out low-confidence detections \cite{liPolyMOTPolyhedralFramework2023}, but selecting the appropriate SF threshold is challenging as it is difficult to control the ratio of true positives (TP) to false positives (FP) in the detection process. To address these challenges, a Distance-Based Score Enhancement (DBSE) mechanism is proposed, leveraging distance to enhance the expressiveness of point cloud detection scores and thereby improving the filtering performance of SF.

\textbf{Prediction.} In the prediction phase, the target's position, rotation angle, and other motion characteristics for the next frame are predicted using either  filter-based methods or  learning-based methods \cite{weng3DMultiObjectTracking2020}. Filter-based methods typically employ filters along with corresponding motion models to adapt to the target's motion patterns and predict its future state. There are two main issues: (1) Different types of targets exhibit distinct motion patterns. (2) The motion patterns of the same target may vary at different stages of its movement. To tackle the first issue, Poly-MOT introduces a classification-based multi-model filter, which effectively predicts and tracks targets of different categories \cite{240313443FastPolyFast,liPolyMOTPolyhedralFramework2023}. The existence of issue (2) leads to model mismatch in tracking the motion of a single target.
Fig. \ref{fig1} illustrates the continuous process of a car making a left turn. Throughout this process, the target transitions through three distinct motion models, with two instances of model switching. It is evident that using a single motion model for tracking results in delays in model adaptation when the target's motion pattern changes. To overcome this limitation, we introduce an interacting multiple model (IMM) algorithm for tracking prediction, which can simultaneously address both the variation in motion patterns across different targets and the changes in motion behavior within a single target's trajectory.

\textbf{Trajectory Manager.} The trajectory management module plays a crucial role in determining the creation and termination of trajectories, which directly influences the number of FP and false negatives (FN). In count-based methods \cite{liPolyMOTPolyhedralFramework2023,benbarkaScoreRefinementConfidencebased2021}, the validity of a trajectory is assessed based on the number of consecutive associations or non-associations. However, this approach heavily relies on parameter settings, and the criteria for validity are relatively simplistic.Another method involves iteratively calculating trajectory confidence and comparing it to a threshold to decide whether a trajectory should be initiated or terminated \cite{240313443FastPolyFast,benbarkaScoreRefinementConfidencebased2021}. However, this approach can easily result in the termination of low-confidence true targets, especially those at the detection threshold, due to the natural decay of trajectory scores. To address this issue, we propose a damping window-based trajectory management method, which incorporates the trajectory's association status and uses a damping window score to evaluate the trajectory's validity. This approach helps mitigate the risk of overlooking low-confidence true detections.

\section{The Proposed Method}
IMM-MOT can be divided into four components: the preprocessing module with a Distance-Based Score Enhancement (DBSE) mechanism, the Interacting Multiple Model (IMM) tracker, the association module, and the Damping Window (DW) lifecycle management module,  as shown in Fig. \ref{fig3}. We enhance the original Score Filter (SF) and  Non-Maximum Suppression (NMS) modules by incorporating  Distance-Based Score Enhancement  as a preliminary step in the preprocessing module. The IMM filter is used to predict the next frame positions of target trajectories in the prediction module. The  Damping Window  mechanism is adopted to update trajectory states in the trajectory lifecycle management module.

\subsection{Interacting Multi-Model Filter}
In the prediction phase, many previous methods have employed a single motion model for the predictions in various target categories \cite{kimEagerMOT3DMultiObject2021, papaisSWTrackMultipleHypothesis2024}. Poly-MOT \cite{liPolyMOTPolyhedralFramework2023} innovatively introduced a multi-category trajectory motion module, which classifies multi-category targets into two categories according to their motion features. Utilizing different models for predictions in each category, the module demonstrated enhanced effectiveness. However, despite the differences in motion patterns between different target categories, the same target can change motion patterns over its tracking duration. Therefore, if the tracker can adaptively switch between appropriate motion models in real-time throughout the entire tracking process,  it will better accommodate the target's motion characteristics. Thus, our work introduces the IMM algorithm into the tracking framework, thereby enhancing tracking performance for multi-category targets and those exhibiting diverse motion patterns, improving both accuracy and stability in tracking.

The IMM algorithm is the mainstream algorithm in the current multi-model theory. It estimates the state of maneuvering targets by transferring models and combining multiple models. First, the filters of all models in the model set are used for parallel filtering. Then, the model probability at the current moment is calculated based on the residual output of each filter and the prior probability of each model. Finally, the system's measurement data are obtained through a weighted combination of these outputs \cite{yunitaErrorPerformanceAnalysis2020}.

Based on the concept and primary steps of the IMM algorithm \cite{liPerformancePredictionInteracting1993}, and considering the tasks of the 3D MOT prediction phase, we designed the IMM prediction tracker, whose main algorithmic flow is shown in Algorithm \ref{algo1}.

\begin{algorithm}[t]
    \renewcommand{\algorithmicrequire}{\textbf{Input:}}
    \renewcommand{\algorithmicensure}{\textbf{Output:}}
    \caption{Interacting Multiple Model Tracker}
    \label{algo1}
    \begin{algorithmic}[1]
        \REQUIRE
            $X_{t-1}$: State at time t-1,
            $Z_t$: Measurement associated at time t 
        \ENSURE 
            $\hat{X}_t$: Prediction at time $t$
        \STATE Initialize the filters $i$ in the model set, Markov matrix $M$ ($\pi_{ji} \in M$), and model probabilities $\mu_i$, $i, j \in [1, N]$
        \REPEAT
            \STATE Based on $X_{t-1}$, predict the target state $\hat{X}_t^i$ for all models $i$ in parallel, $i \in [1, N]$
            \STATE Compute the fused prediction: $\hat{X}_t = \sum \mu_{t-1}^i \hat{X}_t^i$
            \STATE Based on the associated measurement $Z_t$, compute the residual $S_t^i$ and likelihood $\Lambda_t^i$
            \STATE Update model probabilities: $\mu_t^i = \frac{c_i \Lambda_t^i}{\sum_{j=1}^N c_j \Lambda_t^j}$, where $c_i = \sum_{j=1}^N \pi_{ji} \mu_{k-1}^j$
        \UNTIL{No Input}
    \end{algorithmic}
\end{algorithm}

Step 3 and step 4 represent the prediction phase, while step 5 and step 6 correspond to the update phase, with $N$ denoting the number of models in the model set. In the prediction phase, each filter performs parallel predictions, and $\hat{X}_t$ is obtained through weighted fusion before passing it to the association phase. The successfully associated detections $D_t$ are used for updates, where $D_t$'s measurement data $Z_t$ are used to calculate the residual $S_t^i$ and likelihood $\Lambda_t^i$ for each model in the update phase. This allows the proportion of each model in the model set to be dynamically adjusted according to the target’s motion patterns, thereby more accurately fitting the target’s motion and achieving precise tracking.

In this work, our model set is: \{CV, CA, CTRA, CTRV\}. The Constant Velocity (CV) and Constant Acceleration (CA) models assume straight-line target motion with linear state transition equations and do not account for any deviations. The Constant Turn Rate and Velocity (CTRV) and Constant Turn Rate and Acceleration (CTRA) models extend the previous two models by considering rotation around the z-axis. For the first two models, which are linear models, we directly use an IMM algorithm based on the Kalman filter. For the latter two, which are nonlinear models, we employ an Extended Kalman Filter(EKF) to implement the IMM algorithm. Specifically, based on the traditional IMM algorithm using Kalman filters \cite{liPerformancePredictionInteracting1993}, the prediction of state $x$ and the covariance matrix $P$, as well as the observation model, should follow  (\ref{eq1}), (\ref{eq2}),  (\ref{eq3}) respectively.

\begin{equation}
    \label{eq1}
    x_{k}=f(x_{k-1})  + w_{k-1}.
\end{equation}
\begin{equation}
    \label{eq2}
    P_{t,t-1}=F_tP_{t-1}F_t^T+Q. 
\end{equation}
\begin{equation}
    \label{eq3}
    z_{k}=h(x_{k}) +\nu_{k}.
\end{equation}

The above four models can effectively cover a wide range of target motion scenarios, such as acceleration, constant velocity, and turning. Since the core idea of IMM is weighted fusion, we account for the differences in state vectors between the models and unify the state vectors of the four models. The results are shown in Table \ref{tab1}, where $x$, $y$, $z$ represent the target center position, $w$, $l$, $h$ represent the target's shape information, $v$, $a$ represent velocity and acceleration, and $\theta$, $\omega$ represent the heading angle and turning rate of the target, respectively.
\begin{table}[]
    \centering
    \caption{Unified State Vectors in IMM Tracker Models}
    \label{tab1}
    \begin{tabular}{cc p{0.7\linewidth}}
    \toprule
         Model   &  State Vector \\
    \midrule
         CV    &   ${[ x, \, \, y, \,  \, z,  \, \, w, \, \, l, \, \, h, \, \, v_x, \, \, v_y, \, \, v_z, \, \, \theta]}$ \\
         CA    &   ${[ x,\, \,y,\, \,z,\, \,w,\, \,l,\, \,h,\, \,v_x,\, \,v_y,\, \,v_z,\, \,a_x,\, \,a_y,\, \,a_z,\, \,\theta]}$\\
         CTRV  &   ${[ x,\, \,y,\, \,z,\, \,w,\, \,l,\, \,h,\, \,v,\, \,\theta,\, \,\omega]}$\\
         CTRA  &   ${[ x,\, \,y,\, \,z,\, \,w,\, \,l,\, \,h,\, \,v,\, \,a,\, \,\theta,\, \,\omega]}$\\
         IMM   &   ${[ x,\, \,y,\, \,z,\, \,w,\, \,l,\, \,h,\, \,v_x,\, \,v_y,\, \,v_z,\, \,a_x,\, \,a_y,\, \,a_z,\, \, \theta, \, \, \omega ]}$ \\
    \bottomrule
    \end{tabular}
\end{table}

\subsection{Damping Window Trajectory Management}
As for trajectory validity, the following is considered:  
(1) The detections closer to the current time have a greater impact.
(2) The association's information conveyed by consecutive successes, consecutive failures, and intermittent associations is different.  
(3) We should consider all detections(predictions) of the trajectory as a whole when calculating the trajectory score, reducing the impact of score fluctuations in the trajectory.  
Based on these principles, we design the trajectory score update function, as shown in (\ref{eq4}).
We assume that $D$ represents a detection(or prediction) of a trajectory while a trajectory can be represented by $T = \{D_0,D_1,...,D_{k}\}$.

\begin{equation}
    \label{eq4}
    s(t) = \frac{\sum_{i=0}^{k}w_i \cdot f(i-t)}{\sum_{i=0}^{k}f(i-t)}. 
\end{equation}

The damping function \( f(x) \) varies with time, and several constraints apply to it (\ref{eq5},  \ref{eq6}, \ref{eq7}, \ref{eq8}):
\begin{equation}
    \label{eq5}
    f(x) > 0 , x \in (-\infty,0]. 
\end{equation}
\begin{equation}
 \label{eq6}
    \lim_{x \to -\infty} f(x) = 0.
\end{equation}
\begin{equation}
    \label{eq7}
    \frac{df(x)}{x} > 0,x \in (-\infty,0].
\end{equation}
\begin{equation}
    \label{eq8}
    f(0) = c,c>0.
\end{equation}

We take the time lag \( \Delta t \) between each detection in the trajectory and the current time as the independent variable, with the value of \( f(\Delta t) \) as the dependent variable. Considering the temporal factor ($\Delta t$ for all detections  move toward \( -\infty \) over time), we select a function defined on \( (-\infty, 0] \). Actually, \( f(x) \) represents the weights of detections in the trajectory at a specific moment, where detections farther from the current time (as $x$ approaches \( -\infty \)) receive lower weights. We require that as the variable $x$ approaches \( -\infty \), the value of $f(x)$ asymptotically damps to 0. Given the temporal factor, we want higher weights for detections closer to the current time,  while  \( f(0) \) is greater than 0.

In the expression, \( w_i \) denotes the association factor for each point in the trajectory, which is assigned as shown in (\ref{eq9}).

\begin{equation}
\label{eq9}
w_i = 
\left\{
\begin{aligned} 
1&,  \ \ \ \  \text{if }\ associated,\\ 
0&,  \ \ \ \  \text{if }\ not \ associated. 
\end{aligned}
\right. 
\end{equation}

Hence, ensuring that:
$0 \leq s(t) \leq 1$.
In fact, the numerator in (\ref{eq4}) represents a convolution, which is often used to compute the accumulated impact of a time-varying system. The trajectory score fits this scenario, accounting for both the temporal evolution of the score and all points in the trajectory. We assume five scenarios with different conditions:
\begin{itemize}
    \item Condition 1: Associated in the first frame, with no further associations.  
    \item Condition 2: Associated in the first three frames, with no further associations.  
    \item Condition 3: Associated in the first frame and then every alternate frame thereafter.  
    \item Condition 4: Associated in the first frame and then every fourth frame thereafter.  
    \item Condition 5: Associated in the first frame and then associated at intervals determined by varying conditions.  
\end{itemize}

\begin{figure}[t]
  \centering
  \resizebox{\columnwidth}{!}{\includegraphics{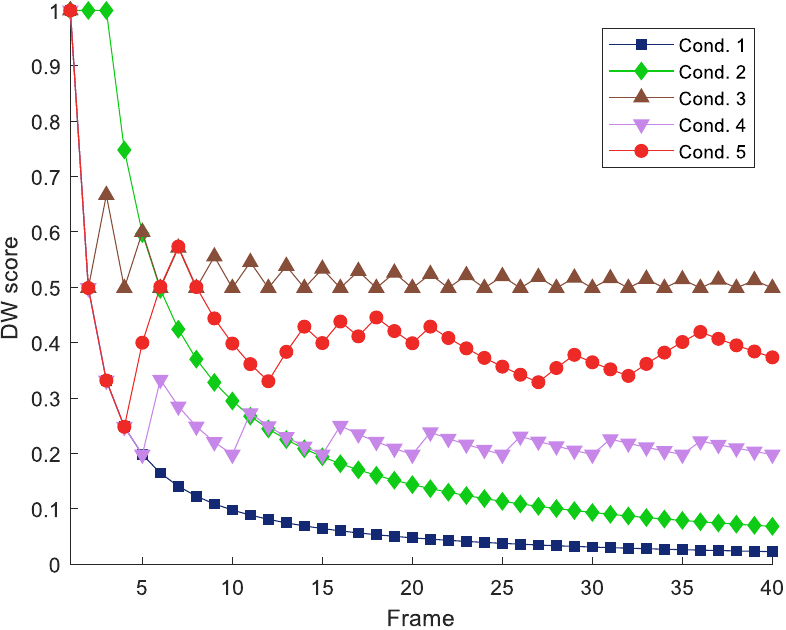}}
  
  \caption{The trend of Damping Window scores for five different conditions.}
  \label{fig4}
\end{figure}

From Fig. \ref{fig4}, it can be observed that if a trajectory is associated multiple times, its DW score fluctuates but remains difficult to terminate, which aligns well with practical scenarios. If a trajectory is not associated multiple times, its score decreases to a certain extent but does not drop too low, thereby preventing the neglect of track fragmentation caused by factors such as occlusion. As shown in Condition 5, random scenarios demonstrate that DW not only ensures penalties for unassociated trajectories but also possesses the ability to tolerate track fragmentation.

\begin{table*}[h]
  \centering
  \caption{The comparison of existing methods and our proposed method on the NuScenes Val dataset. \textbf{The best result is highlighted in \textcolor{red}{red}, and the second-best result is highlighted in \textcolor{blue}{blue}.}}
  \centering
  \label{tab2}
  \begin{tabular}{cccccccc}
    \hline
    Methed & Detector & Input & AMOTA$\uparrow$ & AMOTP$\downarrow$ & IDS$\downarrow$ & FP$\downarrow$ & FN$\downarrow$\\
    \hline
    CBMOT\cite{benbarkaScoreRefinementConfidencebased2021}   & CenterPoint\&CenterTrack   & 2D+3D & 72.0 & \textcolor{red}{48.7}   & 479  & -   & -\\

    EagerMOT\cite{kimEagerMOT3DMultiObject2021}              & CenterPoint\&Cascade R-CNN & 2D+3D & 71.2 & 56.9  & 899   &-    &-\\

    SimpleTrack\cite{pangSimpleTrackUnderstandingRethinking2021} & CenterPoint            & 3D    & 69.6 & 54.7  & 405   &-    &-\\

    CenterPoint\cite{yinCenterbased3DObject2021}             & CenterPoint                & 3D    & 66.5 & 56.7  & 562   &-    &-\\

    OGR3MOT\cite{zaechLearnableOnlineGraph2022}                                                  & CenterPoint                & 3D    & 69.3 & 62.7  & \textcolor{blue}{262}   &-    &-\\

    Poly-MOT\cite{liPolyMOTPolyhedralFramework2023}          & CenterPoint                & 3D    & 73.1 & 51.7  & \textcolor{red}{232} &\textcolor{red}{13051} &17593\\

    Fast-Poly\cite{240313443FastPolyFast}             & CenterPoint                & 3D    & \textcolor{blue}{73.7} & 53.0  & 414 &14713 &\textcolor{blue}{15900}\\
    \hline
    Ours                                                     & CenterPoint                & 3D    & \textcolor{red}{73.8} & \textcolor{blue}{51.6} & 326 &\textcolor{blue}{13433} &\textcolor{red}{15658}\\
    \hline
  \end{tabular}

  \end{table*}

  \begin{table*}
    \centering
    \caption{The performance of the various module combinations in the NuScenes Val dataset.  \textbf{The best result is highlighted in \textcolor{red}{red}, and the second-best result is highlighted in \textcolor{blue}{blue}.}}
    \label{tab3}
    \begin{tabular}{c|ccc|cccccccc}
    \hline
    Seq & DBSE & IMM & DW & AMOTA$\uparrow$ & AMOTP$\downarrow$ &  MT$\uparrow$ & ML$\downarrow$  & TP$\uparrow$ & FP$\downarrow$ & FN$\downarrow$ & IDS$\downarrow$\\
    \hline
        1&      - &     - &     -                      & 73.08                  & 51.68                  &4670                   & 1444                 & 84072 & 13051                  & 17593 & \textcolor{red}{232} \\
        2&  $\checkmark$ &     - &     -               & 73.21                  & \textcolor{red}{51.04} &4746                   & 1331                 & 84654 & \textcolor{red}{12510} & 16942 & 301 \\
        3&      - & $\checkmark$ &     -               & 73.59                  & 51.76                  &4690                   & 1433                 & 84303 & 13039                  & 17342 & \textcolor{blue}{252} \\
        4&      - &     - & $\checkmark$               & 73.26                  & 52.27                  &4773                   & 1292                 & 85161 & \textcolor{blue}{12669} & 16417 & 319 \\
        5&      - & $\checkmark$ & $\checkmark$        & 73.66 & 52.29                  &\textcolor{blue}{4808} & \textcolor{blue}{1252} & \textcolor{blue}{85722} & 13146                 & \textcolor{blue}{15850} & 325 \\
        6&  $\checkmark$ & $\checkmark$ &     -        & \textcolor{blue}{73.72}                  & \textcolor{blue}{51.05}&4705                   & 1427                 & 84521 & 13384                   & 17124 & \textcolor{blue}{252} \\
        7&  $\checkmark$ &     - & $\checkmark$        & 73.34                  & 51.53                  &4783                   & 1295                 & 85234 & 12850 & 16343 & 320 \\
        8&  $\checkmark$ & $\checkmark$ & $\checkmark$ & \textcolor{red}{73.79} & 51.61                  &\textcolor{red}{4827}  & \textcolor{red}{1246}& \textcolor{red}{85913} & 13433  & \textcolor{red}{15658} & 326 \\
        \hline
    \end{tabular}
\end{table*}

\begin{table*}

    \centering
    \caption{The comparison of AMOTA between the IMM module and the baseline for different target types. \textbf{The best result is highlighted in bold.}}
    \label{tab5}
    \begin{tabular}{ccccccccccc}
    \hline
     Set &Method & Detector & Overall & Bic. & Bus & Car & Motor. & Ped. & Tra. & Tru. \\ 
    \hline
     \multirow{2}{*}{Val}& Poly-MOT\cite{liPolyMOTPolyhedralFramework2023} &CenterPoint\cite{yinCenterbased3DObject2021} &  73.1          & 54.6 & 87.3                   & 86.3 & 78.3                   & \textbf{83.0} & 51.0          & 71.2\\
                         &Ours                      &   CenterPoint\cite{yinCenterbased3DObject2021}                    &  \textbf{73.6} & \textbf{56.4} & 87.3 & \textbf{86.4} & \textbf{78.9} & 82.8          & \textbf{51.6} & \textbf{71.6}\\
     \midrule
     \multirow{2}{*}{Test}& Poly-MOT$^*$ & LargeKernel\cite{chenLargeKernel3DScalingKernels2023} &  74.2 & 54.5 & 76.7 & 85.6 & 79.3 & 81.2 & 75.4 & 66.6\\
                        &Ours                    &   LargeKernel\cite{chenLargeKernel3DScalingKernels2023}                      &  \textbf{74.9} & \textbf{54.6} & \textbf{77.0} & \textbf{86.0} & \textbf{79.6} & \textbf{84.1} & \textbf{75.8} & \textbf{67.0}\\
    \bottomrule
    \end{tabular}

    \begin{tablenotes}
        \vspace{1mm}
        \item[1] \parbox[t]{0.6\linewidth}{\ \ \ \ \ \ \ \ \ \ \ \ \ \ \ \ \ \ \ \ * indicates the result we reproduced.}
    \end{tablenotes}

\end{table*}

\subsection{Distance-Based Score Enhancement}
The data obtained during the preprocessing phase are crucial for the entire tracking process. However, we observe that the current detection results for lidar point cloud data overlook an important feature: when lidar detects targets, the point cloud data of nearby targets are denser, while data of distant targets are sparser. Therefore, we argue that detections based on denser point cloud data are more reliable than those based on sparse data. The distribution of noise varies for different targets, as well as for the same target at different distances. Hence, we designed the DBSE mechanism to improve the existing detection scores by incorporating distance factors, reflecting how detection distance impacts the reliability of the target.
We designed a weight function $f(d)$ that decreases with the distance variable to calculate the distance between each detection target and the sensor, yielding a weight for each detection. The enhanced detection score is then computed as shown in (\ref{eq11}).
\begin{figure}
    \centering
    \resizebox{\columnwidth}{!}{\includegraphics{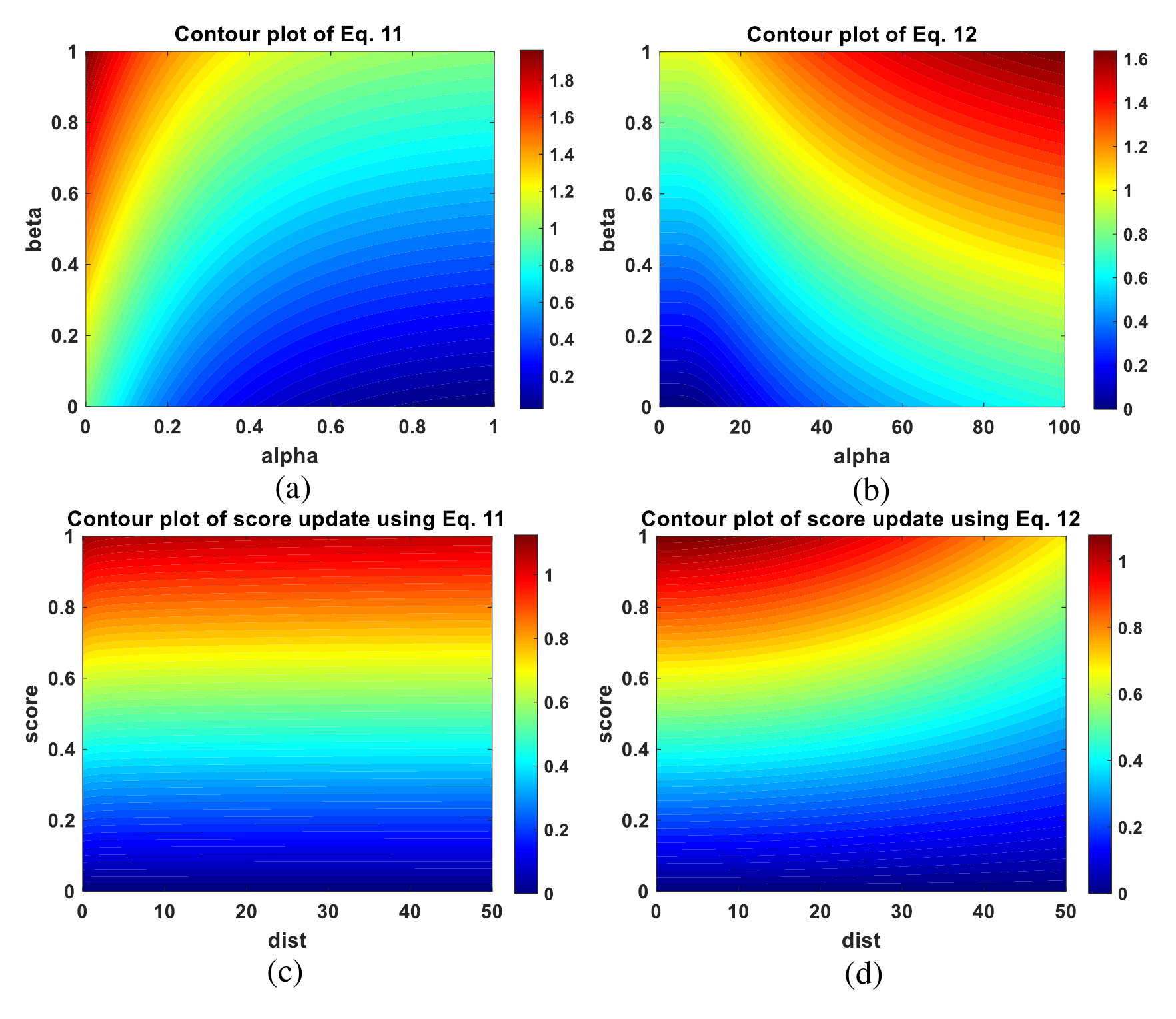}}
    
    \caption{(a) The effect of the parameters in (\ref{eq12}) on the function value. (b) The effect of the parameters in (\ref{eq13}) on the function value. (c) The effect of (\ref{eq12}) on the target score. (d) The effect of (\ref{eq13}) on the target score.}                                                               
    \label{fig5}
\end{figure}

\begin{equation}
    \label{eq11}
    S_{i}^{*} = S_i * f(d_i) 
\end{equation}

Where \( S_i \) represents the detection score for target \( i \), \( d_i \) is the distance between the target and the lidar sensor, and \( S^*_i \) is the enhanced detection score. The following functions (\ref{eq12}), (\ref{eq13}) are designed  for  \( f(d) \):

\begin{equation}
    \label{eq12}
    f(d) = \frac{1}{d^\alpha}+\beta  
\end{equation}
\begin{equation}
    \label{eq13}
    f(d) = e^{-(\frac{d}{\alpha})}+\beta
\end{equation}

Fig. \ref{fig5} (a) and (b) show the impact of parameters $\alpha$ and $\beta$ on the function result when $d = 40$. We control the influence of the function on detection scores at different distances by adjusting the values of $\alpha$ and $\beta$. (c) and (d) illustrate the effect of the two functions on target scores. The values of $\alpha$ and $\beta$ match those used in the experimental section. It can be observed that  (\ref{eq12}) produces a relatively smooth change in scores, making it suitable for targets with strong detection features and good noise resistance. Equation (\ref{eq13})’s design results in a significant impact of distance on the scores, making it suitable for targets that are difficult to detect and sensitive to noise.

\section{Experiment}
\subsection{Dateset}
The NuScenes dataset \cite{caesarNuScenesMultimodalDataset2020} is a large-scale public dataset designed for autonomous driving. It consists of 1,000 driving scenes, each lasting 20 seconds, which include both left-hand and right-hand traffic and cover complex weather conditions, providing a more realistic representation of autonomous driving environments. Additionally, keyframes are sampled at a frequency of 2Hz for each scene, and true bounding box annotations are provided for 40 frames within a single scene. We validate the proposed model on the NuScenes dataset.
\subsection{Implementation Details}
In the  Distance-Based Score Enhancement (DBSE) preprocessing module, we use  (\ref{eq11}) for the car and trailer, with ($\alpha$, $\beta$) set to (0.01, 0.1). For the bus and bicycle, we apply (\ref{eq12}), with ($\alpha$, $\beta$) set to (70, 0.1), (90, 0.2) respectively. The Score Filter (SF) and  Non-Maximum Suppression (NMS) parameters are configured in the same manner as in Poly-MOT \cite{liPolyMOTPolyhedralFramework2023}. In the Interacting Multiple Model (IMM) experiment, the IMM filter is applied to targets from five classes: bicycle, bus, car, motorcycle, and truck. The model set used in IMM consists of {CTRA, CTRV, CA, CV}. In the trajectory management module, the Damping Window (DW) mechanism is applied to the bus, car, pedestrian, and trailer. The thresholds for active and tentative states are set as follows: (0.4, 0.05) for bus, (0.3, 0.05) for car, (0.3, 0.1) for pedestrian, and (0.6, 0.1) for trailer.
\subsection{Comparative Evaluation}
Our method is compared with several well-known, previously published 3D MOT models on the NuScenes Val dataset, including two prominent multi-modal models, CBMOT and EagerMOT, as well as several LiDAR-only models \cite{kimEagerMOT3DMultiObject2021,benbarkaScoreRefinementConfidencebased2021}. The point cloud detector CenterPoint's result is used as input to ensure fairness in the comparison.

As shown in Table \ref{tab2}, IMM-MOT's AMOTA has surpassed the previous highest value achieved by Fast-Poly, reaching an impressive 73.8\%, demonstrating strong performance. Additionally, AMOTP of our method exceeds that of all listed single-modal models, second only to the multi-modal model CBMOT. This is due to the use of the IMM filter as the trajectory predictor, which significantly enhances the accuracy of the tracking process. Furthermore, the reduction in the number of false negatives (FN) achieved by IMM-MOT is noteworthy, which can be attributed to the thoughtful design of the DBSE and DW mechanisms.

\begin{figure}[t]
  \centering
  \resizebox{\columnwidth}{!}{\includegraphics{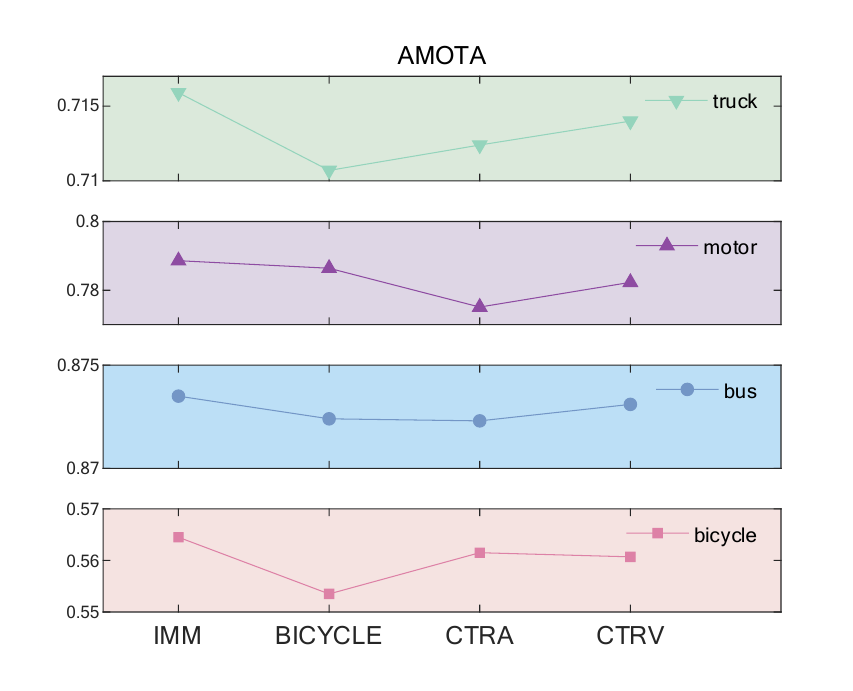}}
  
  \caption{The comparison of Interacting Multiple Model and other models on different targets.}
  \label{fig6}
\end{figure}

\subsection{Ablation Studies}
\textbf{Combinations. }The performance of proposed module combinations is tested in Table \ref{tab3}. As we can see, the IMM module contributes the most significant improvement in AMOTA, enhancing tracking accuracy while maintaining a low number of Identity Switches (IDS). The DBSE plays a crucial role in reducing the number of false positives (FP) by -4\%, which aligns with the goal of minimizing FP and increasing true positives (TP). Meanwhile, the DW mechanism contributes to reductions in FP (-3\%) and FN (-7\%), with a particular improvement in FN, demonstrating better performance compared to the baseline. Additionally, it is notable that although individual metrics did not reach their highest values in the two-module combinations, the overall performance in terms of AMOTA and AMOTP shows that each individual module contributes significantly to the overall performance. This indicates that our method has high compatibility. In the combination of all modules, the results for AMOTA, MT (Most Tracked), ML (Most Losses), TP, and FN all reached their best values, which is consistent with expectations.

\begin{table}[t]
    \centering
    \caption{The experimental performance of the damping window module across different target types.\textbf{The best result is highlighted in bold.}
    \label{tab7}}
    \resizebox{\linewidth}{!}{
        \begin{tabular}{cccccccc}
        \toprule
         & DW & AMOTA$\uparrow$ & MOTA$\uparrow$ & TP$\uparrow$ & FP$\downarrow$ & FN$\downarrow$ & $X_1\downarrow$ \\
        \midrule
        \multirow{2}{*}{Car} & $\checkmark$ & \textbf{86.8} & \textbf{75.8} & \textbf{50351} & \textbf{6154} & \textbf{7849} & \textbf{13.5} \\
                             &      -       & 86.3          & 73.5          & 49064          & 6189          & 9177          & 15.8 \\
        \midrule
        \multirow{2}{*}{Trailer} & $\checkmark$ & \textbf{51.3} & \textbf{41.3} & \textbf{1423} & 421 & \textbf{1001} & \textbf{41.3} \\
                                 &     -        & 51.0          & 38.8          & 1362          & 421 & 1062          & 43.8 \\
        \bottomrule
    \end{tabular}
    }
\end{table}

\begin{table}[t]
    \centering
    \caption{The comparison between the damping window module and other tracking methods. \textbf{The best result is highlighted in bold.}}
    \label{tab8}
    \resizebox{\linewidth}{!}{
    \begin{tabular}{ccccccc}
    \toprule
    Method & AMOTA$\uparrow$ & MOTA$\uparrow$ & TP$\uparrow$ & FP$\downarrow$ & FN$\downarrow$ &IDS$\downarrow$ \\
    \midrule
    DW     & \textbf{73.8}  & \textbf{62.9} & \textbf{85913} & 13433 & \textbf{15658} & 326 \\
    MaxAge\&Confidence & 73.7  & 62.2 & 84521 & \textbf{13384} & 17124 & \textbf{252} \\
    MaxAge & 72.1  & 61.1 & 85797 & 13676 & 15706 & 340 \\
    \bottomrule
    \end{tabular}
    }
\end{table}

\begin{table}[t]
    \centering
    \caption{The experimental performance of the Distance-Based Score Enhancement module across different target types. \textbf{The best result is highlighted in bold.}}
    \label{tab4}
    \resizebox{\linewidth}{!}{
    \begin{tabular}{cccccccc}
        \toprule
        & DBSE & AMOTA$\uparrow$ & MOTA$\uparrow$ & TP$\uparrow$ & FP$\downarrow$ & FN$\downarrow$ & $X_2\downarrow$ \\
        \midrule
        \multirow{2}{*}{Bicycle} & $\checkmark$ & \textbf{54.6} & 48.0 & 1203          & \textbf{246} & 790           &  \textbf{12.3} \\
                                 &     -        & 54.5 & \textbf{48.7} & \textbf{1245} & 275          & \textbf{746}  & 13.8 \\
        \midrule
        \multirow{2}{*}{Trailer} & $\checkmark$ & 50.9          & \textbf{39.6} & \textbf{1372} & \textbf{412} & \textbf{1052}  & \textbf{17.0} \\
                                 &     -        & \textbf{51.0} & 38.8          & 1362          & 421          & 1062           & 17.4 \\
        \bottomrule
    \end{tabular}
    }
\end{table}

\textbf{The effect of the IMM}. (1) On the NuScenes Val Set, we replaced the original Multi-Category Trajectory Motion Module of baseline with the IMM prediction module, as shown in Table \ref{tab5}, resulting in a +0.5\% performance improvement. To validate its tracking performance, we tested four representative target types (bicycle, bus, motorcycle, and truck) and compared the IMM tracker with the Bicycle, CTRA, and CTRV models in Fig. \ref{fig5}. The IMM tracker demonstrated stronger performance on targets with complex motion patterns, surpassing the other models across all tested types. (2) On the Test Set in Table \ref{tab5}, the IMM tracker improved performance by +0.7\%, surpassing the original prediction module and achieving better results across various target categories, further demonstrating its accuracy and stability.

\textbf{The effect of the DW}. As shown in Table \ref{tab7}, we compared the tracking performance using the DW mechanism. DW played a significant role in reducing FN, especially for the car (-14.5\%). Furthermore, while increasing the number of TP, DW also reduced the FN/(TP+FN) ratio, which reflects the proportion of false negative detections caused by the decision-making mechanism during the tracking process. This ensures that more low-confidence edge detections are not overlooked. Additionally, we compared the DW mechanism with other trajectory management methods in Table \ref{tab8}.  Compared to the other methods, DW significantly reduced the proportion of FN while maintaining high accuracy, leading to an overall improvement in tracking quality. In the table, $X_1$ represents
$
\frac{FN}{TP+FN}. 
$

\textbf{The effect of the DBSE}. We compared the tracking performance using DBSE in Table \ref{tab4}. As observed, DBSE not only reduced FP, but also decreased the ratio between FP and the number of true targets, thereby improving the quality of the detections. This aligns with the original design intent of DBSE. In the table, $X_2$ represents
$
\frac{FP}{TP+FN}.
$

\section{Conclusion}
In this work, we propose a new 3D MOT framework: IMM-MOT. Building on the baseline, three novel mechanisms are proposed to achieve effective input data filtering, precise prediction, and more comprehensive trajectory management, including: (1) Interacting Multiple Model (IMM) Filter combines multiple models to predict the complex motion patterns of individual targets, which significantly enhances tracking accuracy. (2) The Damping Window (DW) mechanism controls the birth and termination of trajectories based on their association status, ensuring that low-confidence true detections are no longer overlooked. (3) Incorporating distance to enhance detection scores, the  Distance-Based Score Enhancement (DBSE) mechanism improves the distinction between false positives (FP) and true positives (TP). On the NuScenes Val dataset, IMM-MOT outperforms other point cloud single-modal models, achieving an AMOTA of 73.8\%.

While our method demonstrates strong performance, it has limitations such as the DBSE function selection and parameter tuning relying on the detector, and the IMM transition matrix and initial model probabilities being set empirically. Future work will focus on optimizing the tracking model set and minimizing parameter adjustments. This work is open source to inspire further advancements in the field.

\section*{ACKNOWLEDGMENT}
This work was funded in part by the Natural Science Basic Research Program of Shaanxi, China, Grant 2024JC-ZDXM-40, the Innovation Capability Support Program of Shaanxi, China, Grant 2023-CX-TD-08, and the Key Research and Development Program of Shaanxi Province, China, 2024PT-ZCK-18.

\bibliographystyle{IEEEtran}
\bibliography{IEEEabrv,reference}

\end{document}